\documentclass{article} 

\usepackage{graphicx}
\usepackage[english]{babel}
\usepackage{amsmath}
\usepackage{color}
\usepackage{float}
\usepackage [all]{xy}
\usepackage{fancybox}
\usepackage{subfigure}
\usepackage{url}
\usepackage{multicol}
\usepackage[T1]{fontenc}
\usepackage[utf8]{inputenc}
\usepackage{authblk}
\usepackage{amsmath}
\usepackage{amssymb}

\providecommand{\keywords}[1]{\textbf{\textit{Keywords:}} #1}
\begin{document}

\title{A mass-flow MILP formulation for energy-efficient supplying in assembly lines
\thanks{This work was supported by the ECO-INNOVERA-1rst call EASY (ANR-12-INOV-0002).}
}



\author[1]{Maria Muguerza}
\author[1]{Cyril Briand \thanks{briand@laas.fr}}
\author[1]{Nicolas Jozefowiez\thanks{njozefow@laas.fr}}
\author[1]{Sandra Ulrich Ngueveu\thanks{ngueveu@laas.fr}}
\author[2]{Victoria Rodr\'iguez\thanks{vrodriguez@unav.es}}
\author[3]{Matias Urenda Moris\thanks{matias.urenda.moris@his.se}}
\affil[1]{CNRS, LAAS, UPS, INSA, INP, 7 avenue du colonel Roche, F-31400 Toulouse, France }
\affil[2]{Economics and Management School, University of Navarra,, 31080 Pamplona, Spain}
\affil[3]{Virtual Systems Research Centre, University of Skövde, PO Box 408, 54128 Skövde, Sweden}

\renewcommand\Authands{ and }


\maketitle

\begin{abstract}
This paper focuses on the problem of supplying the workstations of assembly lines with components during the production process. For that specific problem, this paper presents a Mixed Integer Linear Program (MILP) that aims at minimizing the energy consumption of the supplying strategy.  More specifically, in contrast of the usual formulations that only consider component flows, this MILP handles the mass flow that are routed from one workstation to the other. \\ 
\end{abstract}

\section{Introduction}
\label{intro}


In general, feeding systems of assembly lines are composed by a central warehouse, several workstations organized in sequence and a fleet of vehicles (tow trains) in charge of delivering the components to the workstations. The components are packed in pallet or boxes. The supermarket is a decentralized area of material supplies, located next to the assembly line. For building up a supplying strategy, time is discretized in a set of delivering periods. For each period, a workstation has a component consumption (possibly periodic) expressed in terms of boxes. At each tour, the tow trains load the boxes which have to be shipped to the assembly line, follow a supplying route, and stop at the appropriate workstations for delivering its boxes. The supplying routes are usually fixed and start and finish at the supermarket. The number of boxes that a tow train can transport in the same tour is limited. The number of boxes available at each workstation should never exceed the storage capacity of the workstation (which is usually low). A supplying strategy defines whether a vehicle has to stop at each workstation at each time period and the number of boxes that should be delivered. 

In a world where natural resources are limited, issues related to energy efficiency are becoming more and more important. Vehicles in factories travel a significant quantity of kilometers for supplying the workstations, causing effects in economic and energetic expenses. Whether they use electric energy or fossil fuel, their energetic consumption is not negligible  and more and more attention has to be paid for exhibiting energy-efficient supplying strategies. Several factors that are inherent to the problem have impact on the energy consumption. We are interested in determining the most significant ones.  


In the literature related to the Vehicle Routing Problem (VRP), some researchers take interest in minimizing carbon dioxide emissions. One of the major contribution is due to Bektas and Laporte~\cite{PRP} who present the Pollution-Routing Problem (PRP) as an extension of the VRP with Time Windows. The PRP consists of routing a number of vehicles to serve a set of customers within preset time windows, and determining their speed on each route segment, so as to minimize a function comprising emissions and driver costs. The author propose a MILP formulation that allows to optimize both load and speed of the vehicles.The idea of controlling the vehicle velocity on each route segment is fruitful for improving the energy efficiency in the context of long distance transportation problem. However, in a very local transportation context, as the distances travelled during the acceleration phase becomes non-negligible with respect to the one covered at the maximum speed, other parameters can impact the energy consumption.

The problem considered in this paper can be viewed as a particular Inventory Routing Problem (IRP) . We intend to show that minimizing the travelled distance does not necessarily implies the minimization of the energy.  We prove that other parameters can significantly  influence the energy spending. The remainder of this paper is organized as follows. An energy consumption analysis is proposed in Section~\ref{sec:energyanalisis}. A MILP for energy optimization is described in Section~\ref{sec:model}.

\section{Energy modeling}
\label{sec:energyanalisis}

The forces that have more influence on the power consumed by the vehicle are: the traction force ($F_{t} = m_{T}a(t)$) and the rolling resistance ($F_{r}m_{T} =g  C_{r}$), in Newtons (N). The traction force is used to generate motion between an object and a tangential surface, and it depends on the mass ($ m_{T}$) and the acceleration of the vehicle ($ a(t)$). The rolling resistance is the force resisting the motion when a body rolls on a surface and varies in function of the load ($ m_{T}$), the rolling coefficient ($C_{r}$) and the gravity ($g$). The parameter ($ m_{T}$) represents the mass of the vehicle plus the transported load, which varies along the tour. The expresion of the energy consumption is $ E = \int { m_{T} (a(t)+ gC_{r}) v(t)dt}$.

 For sake of simplicity, the acceleration, the deceleration  and the maximum speed are assumed known and constant.Thanks to the literature, the rolling coefficient is also known. Regarding the energy consumed between two workstations. Three phases are distinguished according to the vehicle state. The first phase corresponds to the acceleration phase where a peak of energy is produced, due to the acceleration. The second phase begins when the speed of the vehicle reach its maximum value. Finally, in the deceleration phase, the energy consumption is null.

The travelled distance is directly linked to the energy although it is not the only significant parameter. Indeed, the energy consumption is different depending on the way of delivering the load. The stops at the workstations also have effects on the energy demand. Decreasing the number of vehicle stops in every workstation can reduces the number of acceleration phases, hence the energy. 

\section{Energy-aware mathematical modeling}
\label{sec:model}

In this section, a mixed integer linear programming (MILP) model is presented. This model integrates the previous influential factors, and similarly to the formulation poposed in~\cite{IRP}, takes advantage from a basic flow formulation. Nonetheless, instead of taking the flow in terms of number of components into account, the mass of the shipped components is considered. We assume that only one kind of pallet can be delivered to a given workstation, each having a well-known mass. Therefore, once the mass of delivered components known,  the number of components can be easily deduced. Reasoning in terms of masses is interesting since the energy spent for bringing a pallet to one location \textit{i} to another location \textit{j} is proportional to its mass. Therefore, one can considered directly inside the MILP formulation the energy cost ($C_{ij}$), which represents the energy consumption for shipping one mass unit  directly from \textit{i} to \textit{j} with $j>i$.

The component mass brought from \textit{i} to \textit{j} during period \textit{t} is noted $M_{ij}^{t}$. Decision variables $Z_{i}^{t}$ represent the number of components left at workstation \textit{i} during period \textit{t}. They can easily be deduced from the values of the $M_{ij}^{t}$ variables. Eventually, inventory flow variables $IL_{i}^{t}$, deduced from the $Z_{i}^{t}$ values, are also modelled.



Using the above decision variables, the energy minimization MILP can be formulated as follows. The objective function~\eqref{objmasas} aims at minimizing the energy consumption, which is proportional to the mass $M_{ij}^{t}$ traversing each arc (\textit{i,j}) during period $t$. Constraints~\eqref{demandmas} model flow conservation together with demand satisfaction. Constraints~\eqref{capA} ensure that the vehicle capacity $A$ is never exceeded and enforce variables $Y^{t}$ to be set to one when a tour is carried out in period \textit{t}. The set of equations~\eqref{capestacion} ensures that the inventory level at workstation \textit{i} never exceeds the workstation storage capacity $c_i$. Constraints~\eqref{tour} enforce the difference $\frac{1}{m_{i}}(\sum_{j<i}M_{j,i}^{t} - \sum_{j>i}M_{i,j}^{t})$ to be integral. The constraints~\eqref{mv} impose that the mass brought back to the depot equals the vehicle mass ($0$ and $n+1$ being two virtual nodes associated with the depot). Constraints~\eqref{para} ensure that, whether some components are delivered in workstation \textit{i} during period \textit{t}, the vehicle has to stop in this station at that tour. Set of constraints~\eqref{arcos} ensure that, whether the vehicle stops in workstation \textit{i} at time \textit{t}, there exist an incoming and an outcoming arc selected at workstation \textit{i} during period \textit{t}. Constraints~\eqref{masaarco} ensure that whether there exists a mass flow between two workstations, an arc between these stations has to be selected too. Equations~\eqref{dominio2}-\eqref{dominio3} define the domain of each variable ($m_{\max}$ is the maximum load that the vehicle can transport).

\begin{align}
&\textrm{Min} \hspace{0.3cm } z = \hspace{0.3cm} \sum_{i,j}^{n}\sum_{t}^{NT}C_{ij}M_{ij}^{t}& \label{objmasas}
\end{align}

st:
\begin{align}
\label{demandmas}
&Z_{i}^{t} + IL_{i}^{t-1}-d_{i}^{t} = IL_{i}^{t} \hspace{0.3cm }&\forall\hspace{0.15cm } (i, t)\\\label{capA}
&\sum_{i=1}^{n}Z_{i}^{t}\leq AY^{t} \hspace{0.3cm }&\forall\hspace{0.15cm } (t)\\\label{capestacion}
&Z_{i}^{t} + IL_{i}^{t-1} \leq c_{i} \hspace{0.3cm }&\forall\hspace{0.15cm } (i, t)\\\label{tour}
&Z_{i}^{t} - \frac{1}{m_{i}}(\sum_{j<i}M_{ji}^{t} - \sum_{j>i}M_{ij}^{t}) = 0 \hspace{0.3cm }&\forall\hspace{0.15cm } (i, t)\\\label{mv}
&\sum_{i=1}^{n}M_{in+1}^{t}= m_vY^{t} \hspace{0.3cm }&\forall\hspace{0.15cm } (t)\\\label{para}
&Z_{i}^{t}\leq X_{i}^{t}c_{i} \hspace{0.3cm }&\forall\hspace{0.15cm } (i,t)\\\label{arcos}
&\sum_{j>i}\phi_{ij}^{t}=\sum_{j<i}\phi_{ji}^{t} = X_{i}^{t} \hspace{0.3cm }&\forall\hspace{0.15cm } (i,t)\\\label{masaarco}
&M_{ij}^{t}\leq  m_{\max}\phi_{ij}^{t} \hspace{0.3cm }&\forall\hspace{0.15cm } (i,j,t)\\\label{dominio2}
&IL_{i}^{t} ,M_{ij}^{t}\geq 0 \hspace{0.3cm }&\forall\hspace{0.15cm } (i,j,t)\\
&Z_{i}^{t}\in \mathbb{N} \hspace{0.3cm }&\forall\hspace{0.15cm } (i, t)\\
\label{dominio3}
&\phi_{ij}^{t}, X_{i}^{t}, Y_{t}\in \left \{0,1\right \}  \hspace{0.3cm }&\forall\hspace{0.15cm } (i,j,t)
\end{align}

\section{Conclusion}\label{sec:conclusion}

We can conclude that taking the transported load, the number of stops and the total travelled distance simultaneously into account is worthy. We propose a MILP formulation that integrates these parameters all together inside the optimization procedure. Nevertheless, the first experiments show that the computational effort required for solving efficiently the model is high. Additional researches are needed in order to boost the optimization procedure using either more compact MILP formulations or more advanced optimization mechanisms such as valid inequalities generation, variable fixing techniques, or decomposition approaches.\\

\end{document}